\documentclass[11pt,letterpaper]{article}
\usepackage{emnlp2017}
\usepackage{times}
\usepackage{latexsym}

\usepackage{standalone}
\usepackage{amsmath}
\usepackage{amsfonts}
\usepackage{url}
\usepackage{color}
\usepackage{multirow}
\usepackage{booktabs} 
\usepackage{subcaption} 

\emnlpfinalcopy

\def\emnlppaperid{***}

\title{Hierarchical Embeddings for Hypernymy Detection and Directionality}

\author{Kim Anh Nguyen, Maximilian K\"oper, 
		Sabine Schulte im Walde, Ngoc Thang Vu \\
	    Institut f\"ur Maschinelle Sprachverarbeitung\\
	    Universit\"at Stuttgart\\
	    Pfaffenwaldring 5B, 70569 Stuttgart, Germany\\
	    {\{\tt nguyenkh,koepermn,schulte,thangvu\}@ims.uni-stuttgart.de}}

\date{}

\begin{document}

\maketitle

\begin{abstract}
  We present a novel neural model $\mathit{HyperVec}$ to learn
  hierarchical embeddings for hypernymy detection and
  directionality. While previous embeddings have shown limitations on
  prototypical hypernyms, $\mathit{HyperVec}$ represents an
  unsupervised measure where embeddings are learned in a specific
  order and capture the hypernym--hyponym distributional
  hierarchy. Moreover, our model is able to generalize over unseen
  hypernymy pairs, when using only small sets of training data, and by
  mapping to other languages. Results on benchmark datasets show that
  $\mathit{HyperVec}$ outperforms both state-of-the-art unsupervised
  measures and embedding models on hypernymy detection and
  directionality, and on predicting graded lexical entailment.
\end{abstract}

\section{Introduction}

Hypernymy represents a major semantic relation and a key organization
principle of semantic memory~\cite{Miller/Fellbaum:91,Murphy:02}. It
is an asymmetric relation between two terms, a hypernym
(superordinate) and a hyponym (subordiate), as in
\textit{animal--bird} and \textit{flower--rose}, where the hyponym
necessarily implies the hypernym, but not vice versa. From a
computational point of view, automatic hypernymy detection is useful
for NLP tasks such as taxonomy creation~\cite{Snow:2006,Navigli:11}, recognizing
textual entailment~\cite{Dagan:13}, and text
generation~\cite{Biran/McKeown:13}, among many others.

Two families of approaches to identify and discriminate hypernyms are
predominent in NLP, both of them relying on word vector
representations.
\textbf{\textit{Distributional count approaches}} make use of either
directionally unsupervised measures or of supervised classification
methods. Unsupervised measures exploit the \textit{distributional
  inclusion hypothesis}~\cite{Geffet/Dagan:05,Geffet/Dagan:09}, or the
\textit{distributional informativeness
  hypothesis}~\cite{Santus:14,Rimell:14}. These measures assign scores
to semantic relation pairs, and hypernymy scores are expected to be
higher than those of other relation pairs. Typically, Average
Precision (AP)~\cite{Kotlerman:10} is applied to rank and distinguish
between the predicted relations.
Supervised classification methods represent each pair of words as a
single vector, by using the concatenation or the element-wise
difference of their vectors~\cite{Baroni:12,Roller:14,Weeds:14}. The
resulting vector is fed into a Support Vector Machine (SVM) or into
Logistic Regression (LR), to predict hypernymy. Across approaches,
\newcite{Shwartz:17} demonstrated that there is no single unsupervised
measure which consistently deals well with discriminating hypernymy
from other semantic relations. Furthermore, \newcite{Levy:15} showed
that supervised methods memorize \textit{prototypical hypernyms} instead of \textit{learning} a relation between two words.

\textbf{\textit{Approaches of hypernymy-specific embeddings}} utilize
neural models to learn vector representations for
hypernymy. \newcite{Yu:15} proposed a supervised method to learn term
embeddings for hypernymy identification, based on pre-extracted
hypernymy pairs. Recently, \newcite{Tuan:16} proposed a dynamic
weighting neural model to learn term embeddings in which the model
encodes not only the information of hypernyms vs. hyponyms, but also
their contextual information. The performance of this family of models
is typically evaluated by using an SVM to discriminate hypernymy from
other relations.


In this paper, we propose a novel neural model $\mathit{HyperVec}$ to
learn hierarchical embeddings that \textbf{(i)} discriminate hypernymy
from other relations (\textbf{detection task}), and \textbf{(ii)}
distinguish between the hypernym and the hyponym in a given hypernymy
relation pair
(\textbf{directionality task}). Our model learns to strengthen the
distributional similarity of hypernym pairs in comparison to other
relation pairs, by moving hyponym and hypernym vectors close to each
other. In addition, we generate a distributional hierarchy between
hyponyms and hypernyms. Relying on these two new aspects of hypernymy
distributions, the similarity of hypernym pairs receives higher scores
than the similarity of other relation pairs; and the
distributional hierarchy of hyponyms and hypernyms indicates the
directionality of hypernymy.

Our model is inspired by the \textit{distributional inclusion
  hypothesis}, that prominent context words of hyponyms are expected
to appear in a subset of the hypernym contexts. We assume that each
context word which appears with both a hyponym and its hypernym can be
used as an indicator to determine which of the two words is
semantically more general: Common context word vectors which represent
distinctive characteristics of a hyponym are expected to be closer to
the hyponym vector than to its hypernym vector. For example, the
context word \textit{flap} is more characteristic for a \textit{bird}
than for its hypernym \textit{animal}; hence, the vector of
\textit{flap} should be closer to the vector of \textit{bird} than to
the vector of \textit{animal}.

We evaluate our $\mathit{HyperVec}$ model on both unsupervised and
supervised hypernymy detection and directionality tasks. In addition,
we apply the model to the task of graded lexical
entailment~\cite{Vulic:16}, and we assess the capability of
$\mathit{HyperVec}$ on generalizing hypernymy by mapping to German and
Italian. Results on benchmark datasets of hypernymy show that the
hierarchical embeddings outperform state-of-the-art measures and
previous embedding models. Furthermore, the implementation of our models is made publicly available.\footnote{\scriptsize \url{www.ims.uni-stuttgart.de/data/hypervec}}

\section{Related Work}
\label{related-work}

\paragraph{Unsupervised hypernymy measures:}

A variety of directional measures for unsupervised hypernymy
detection~\cite{Weeds/Weir:03,Weeds:04,Clarke:09,Kotlerman:10,Lenci/Benotto:12}
all rely on some variation of the \textit{distributional inclusion
  hypothesis}: If $u$ is a semantically narrower term than $v$, then a
significant number of salient distributional features of $u$ is
expected to be included in the feature vector of $v$ as well. In
addition, \newcite{Santus:14} proposed the \textit{distributional
  informativeness hypothesis}, that hypernyms tend to be less
informative than hyponyms, and that they occur in more general
contexts than their hyponyms.
All of these approaches represent words as vectors in distributional
semantic models~\cite{Turney/Pantel:10}, relying on the
\textit{distributional hypothesis}~\cite{Harris:54,Firth:57}.  For
evaluation, these directional models use the AP measure to assess the
proportion of hypernyms at the top of a score-sorted list.  In a
different vein, \newcite{Kiela:15} introduced three unsupervised
methods drawn from visual properties of images to determine a
concept's generality in hypernymy tasks.

\paragraph{Supervised hypernymy methods:}

The studies in this area are based on word embeddings which represent
words as low-dimensional and real-valued
vectors~\cite{Mikolov:13,Pennington:14}. Each hypernymy pair is
encoded by some combination of the two word vectors, such as
concatenation~\cite{Baroni:12} or
difference~\cite{Roller:14,Weeds:14}. Hypernymy is distinguished from
other relations by using a classification approach, such as SVM or
LR. Because word embeddings are trained for similar and symmetric
vectors, it is however unclear whether the supervised methods do
actually learn the asymmetry in hypernymy~\cite{Levy:15}.

\paragraph{Hypernymy-specific embeddings:}

These approaches are closest to our work. \newcite{Yu:15} proposed a
dynamic distance-margin model to learn term embeddings that capture
properties of hypernymy. The neural model is trained on the taxonomic
relation data which is pre-extracted. The resulting term embeddings
are fed to an SVM classifier to predict hypernymy. However, this model
only learns term pairs without considering their contexts, leading to a
lack of generalization for term embeddings. \newcite{Tuan:16}
introduced a dynamic weighting neural network to learn term embeddings
that encode information about hypernymy and also about their contexts,
considering all words between a hypernym and its hyponym in a
sentence. The proposed model is trained on a set of hypernym relations
extracted from WordNet~\cite{Miller:95}. The embeddings are applied as
features to detect hypernymy, using an SVM
classifier. \newcite{Tuan:16} handles the drawback of the approach by
\newcite{Yu:15}, considering the contextual information between two
terms; however the method still is not able to determine the
directionality of a hypernym pair. \newcite{Vendrov:16} proposed a
method to encode order into learned distributed representations, to
explicitly model partial order structure of the visual-semantic
hierarchy or the hierarchy of hypernymy in WordNet. The resulting
vectors are used to predict the transitive hypernym relations in
WordNet.

\section{Hierarchical Embeddings}
\label{hierarchical-embeddings}

In this section, we present our model of hierarchical embeddings
$\mathit{HyperVec}$. Section~\ref{learning-embeddings} describes how
we learn the embeddings for hypernymy, and Section~\ref{measure}
introduces the unsupervised measure $\mathit{HyperScore}$ that is
applied to the hypernymy tasks.

\subsection{Learning Hierarchical Embeddings}
\label{learning-embeddings}

Our approach makes use of a set of hypernyms which could be obtained 
from either exploiting the transitivity of the hypernymy relation~\cite{Fallucchi/Zanzotto:2011}
or lexical databases, 
to learn hierarchical embeddings. We rely on WordNet, a large lexical
database of English~\cite{Fellbaum:98}, and extract all
hypernym--hyponym pairs for nouns and for verbs, including both direct
and indirect hypernymy, e.g., \textit{animal--bird, bird--robin,
  animal--robin}.
Before training our model, we exclude all hypernym pairs which appear
in any datasets used for evaluation.

In the following, Section~\ref{skip-gram} first describes the
Skip-gram model which is integrated into our model for
optimization. Section~\ref{objective} then describes the objective
functions to train the hierarchical embeddings for hypernymy.


\subsubsection{Skip-gram Model}
\label{skip-gram}

The Skip-gram model is a word embeddings method suggested by
\newcite{Mikolov:13}. \newcite{Levy/Goldberg:14} introduced a variant
of the Skip-gram model with negative sampling (SGNS), in which the
objective function is defined as follows:
\resizebox{0.99\linewidth}{!}{ \centering
  \begin{minipage}{\linewidth}
    \begin{alignat}{2}
      \mathrm{J}_{SGNS} &= &&\sum\limits_{w \in V_W} {\sum\limits_{c \in V_C}} \mathrm{J}_{(w,c)} \label{eq1} \\
      \mathrm{J}_{(w,c)} &= {}&& \#(w,c)\log \sigma (\vec{w},\vec{c}) \notag \\
      &&&+ k \cdot \mathbb{E}_{c_N \sim P_D} [\log \sigma (-\vec{w},\vec{c}_N)] \label{eq2}
    \end{alignat}
    \vspace{0.05em}
  \end{minipage}
} where the skip-gram with negative sampling is trained on a corpus of
words $w \in V_W$ and their contexts $c \in V_C$, with $V_W$ and $V_C$
the word and context vocabularies, respectively. The collection of
observed words and context pairs is denoted as $D$; the term $\#(w,c)$
refers to the number of times the pair $(w,c)$ appeared in $D$; the
term $\sigma(x)$ is the sigmoid function; the term $k$ is the number
of negative samples and the term $c_N$ is the sampled context, drawn
according to the empirical unigram distribution $P$. 

\subsubsection{Hierarchical Hypernymy Model}
\label{objective}

Vector representations for detecting hypernymy are usually encoded by
standard first-order distributional co-occurrences. In this way, they
are insufficient to differentiate hypernymy from other paradigmatic relations such
as synonymy, meronymy, antonymy, etc. Incorporating directional
measures of hypernymy to detect hypernymy by exploiting the common
contexts of hypernym and hyponym improves this relation distinction, but
still suffers from distinguishing between hypernymy and
meronymy.

Our novel approach presents two solutions to deal with these
challenges. First of all, the embeddings are learned in a specific
order, such that the similarity score for hypernymy is higher than the
similarity score for other relations. For example, the hypernym pair
\textit{animal--frog} will be assigned a higher cosine score than the
co-hyponymy pair \textit{eagle--frog}. Secondly, the embeddings are
learned to capture the distributional hierarchy between hyponym and
hypernym, as an indicator to differentiate between hypernym and
hyponym. For example, given a hyponym--hypernym pair ($p,q$), we
can exploit the Euclidean norms of $\vec{q}$ and $\vec{p}$ to
differentiate between the two words, such that the Euclidean norm of
the hypernym $\vec{q}$ is larger than the Euclidean norm of the hyponym
$\vec{p}$.

Inspired by the distributional lexical contrast model in
\newcite{Nguyen:16} for distinguishing antonymy from synonymy, this paper proposes two objective functions to learn hierarchical embeddings for hypernymy. Before moving to the details of the two objective functions, we first define the terms as follows: $\mathbb{W}(c)$ refers to the set of words co-occurring with the context $c$ in a certain window-size; 
$\mathbb{H}(w)$ denotes the set of hypernyms for the word $w$; the two terms $\mathbb{H^+}(w,c)$ and $\mathbb{H^-}(w,c)$ are drawn from $\mathbb{H}(w)$, and are defined as
follows:
\resizebox{0.99\linewidth}{!}{ \centering
  \begin{minipage}{\linewidth}
    \begin{alignat}{2}
      \mathbb{H}^{+}(w,c) &= \{u \in \mathbb{W}(c) \cap \mathbb{H}(w) : cos(\vec{w},\vec{c}) - cos(\vec{u},\vec{c}) \geq \theta\} \notag \\
      \mathbb{H}^{-}(w,c) &= \{v \in \mathbb{W}(c) \cap \mathbb{H}(w) : cos(\vec{w},\vec{c}) - cos(\vec{v},\vec{c}) < \theta\} \notag
    \end{alignat}
    \vspace{0.05em}
  \end{minipage}
}
where $cos(\vec{x},\vec{y})$ stands for the cosine similarity of the two
vectors $\vec{x}$ and $\vec{y}$; $\theta$ is the margin. The set
$\mathbb{H}^{+}(w,c)$ contains all hypernyms of the word $w$ that
share the context $c$ and satisfy the constraint that the
cosine similarity of pair $(w,c)$ is higher than the cosine similarity
of pair $(u,c)$ within a max-margin framework $\theta$. Similarly, the
set $\mathbb{H}^{-}(w,c)$ represents all hypernyms of the word $w$
with respect to the common context $c$ in which the cosine similarity
difference between the pair $(w,c)$ and the pair $(v,c)$ is within a
min-margin framework $\theta$. The two objective functions are defined as follows:
\resizebox{0.99\linewidth}{!}{ \centering
  \begin{minipage}{\linewidth}
    \begin{alignat}{2}
      \mathrm{L}_{(w,c)} &= \frac{1}{\#(w,u)} \sum\nolimits_{u \in \mathbb{H}^{+}(w,c)} {\partial(\vec{w},\vec{u})} \label{eq3} \\
      \mathrm{L}_{(v,w,c)} &= \sum\nolimits_{v \in \mathbb{H}^{-}(w,c)} {\partial(\vec{v},\vec{w})} \label{eq4}
    \end{alignat}
    \vspace{0.05em}
  \end{minipage}
} where the term $\partial(\vec{x},\vec{y})$ stands for the cosine derivative of $(\vec{x},\vec{y})$; and $\partial$ then is optimized by the negative sampling procedure.

The objective function in Equation~\ref{eq3} minimizes the
distributional difference between the hyponym $w$ and the hypernym $u$
by exploiting the common context $c$. More specifically, if the common
context $c$ is the distinctive characteristic of the hyponym $w$
(i.e. the common context $c$ is closer to the hyponym $w$ than to the
hypernym $u$), the objective function $\mathrm{L}_{(w,c)}$ tries to
decrease the distributional generality of hypernym $u$ by moving $w$
closer to $u$. For example, given a hypernym-hyponym pair
\textit{animal--bird}, the context \textit{flap} is a distinctive characteristic of \textit{bird}, because almost every \textit{bird} can flap, but not every \textit{animal} can flap. 
Therefore, the context \textit{flap} is closer to the
hyponym \textit{bird} than to the hypernym \textit{animal}.
The model then tries to move \textit{bird} closer to
\textit{animal} in order to enforce the similarity between
\textit{bird} and \textit{animal}, and to decrease the distributional
generality of \textit{animal}. 

In contrast to Equation~\ref{eq3}, the objective function in
Equation~\ref{eq4} minimizes the distributional difference between the
hyponym $w$ and the hypernym $v$ by exploiting the common context $c$,
which is a distinctive characteristic of the hypernym $v$. In this
case, the objective function $\mathrm{L}_{(v,w,c)}$ tries to reduce
the distributional generality of hyponym $w$ by moving $v$ closer to
$w$. For example, the context word \textit{rights}, a distinctive
characteristic of the hypernym \textit{animal}, should be closer to
\textit{animal} than to \textit{bird}. Hence, the model tries to move
the hypernym \textit{animal} closer to the hyponym
\textit{bird}. Given that hypernymy is an asymmetric and also a
hierarchical relation, where each hypernym may contain several
hyponyms, our objective functions updates simultaneously both the
hypernym and all of its hyponyms; therefore, our objective functions
are able to capture the hierarchical relations between the hypernym
and its hyponyms. Moreover, in our model, the margin framework $\theta$ plays a
role in learning the hierarchy of hypernymy, and in 
preventing the model from minimizing the distance of synonymy
or antonymy, because synonymy and antonymy share many contexts. 

In the final step, the objective function which is used to learn the
hierarchical embeddings for hypernymy combines
Equations~\ref{eq1},~\ref{eq2},~\ref{eq3}, and \ref{eq4} by the
objective function in Equations~\ref{eq5} and \ref{eq6}:
\resizebox{0.99\linewidth}{!}{ \centering
  \begin{minipage}{\linewidth}
    \begin{alignat}{1}
      \mathrm{J}_{(w,v,c)} &= \mathrm{J}_{(w,c)} + \mathrm{L}_{(w,c)} + \mathrm{L}_{(v,w,c)} \label{eq5} \\
      \mathrm{J} &= \sum\limits_{w \in V_W} {\sum\limits_{c \in V_C}} \mathrm{J}_{(w,v,c)} \label{eq6}
    \end{alignat}
    \vspace{0.05em}
  \end{minipage}
}

\subsection{Unsupervised Hypernymy Measure}
\label{measure}

$\mathit{HyperVec}$ is expected to show the two following properties:
(i) the hyponym and the hypernym are close to each other, and (ii)
there exists a distributional hierarchy between hypernyms and their
hyponyms. Given a hypernymy pair $(u,v)$ in which $u$ is the hyponym
and $v$ is the hypernym, we propose a measure to detect hypernymy and
to determine the directionality of hypernymy by using the hierarchical
embeddings as follows:
\begin{equation}
  \mathit{HyperScore}(u,v) = cos(\vec{u},\vec{v}) * \frac{\|\vec{v}\|}{\|\vec{u}\|}
  \label{eq7}
\end{equation}
where $cos(\vec{u},\vec{v})$ is the cosine similarity between
$\vec{u}$ and $\vec{v}$, and $\|\cdot\|$ is the magnitude of the
vector (or the Euclidean norm). The cosine similarity is applied to
distinguish hypernymy from other relations, due to the first property
of the hierarchical embeddings, while the second property is used to
decide about the directionality of hypernymy, assuming that the
magnitude of the hypernym is larger than the magnitude of the
hyponym. Note that the proposed hypernymy measure is unsupervised when the resource is only used to learn hierarchical embeddings.

\section{Experiments}
\label{experiments}

In this section, we first describe the experimental settings in our
experiments (Section~\ref{subsec:settings}). We then evaluate the
performance of $\mathit{HyperVec}$ on three different tasks: i)
unsupervised hypernymy detection and directionality
(Section~\ref{subsec:unsupervised-detection-directionality}), where we
assess $\mathit{HyperVec}$ on ranking and classifying hypernymy; ii)
supervised hypernymy detection
(Section~\ref{subsec:supervised-hypernymy-detection}), where we apply
supervised classification to detect hypernymy; iii) graded lexical
entailment (Section~\ref{subsec:graded-lexical-entailment}), where we
predict the strength of hypernymy pairs.

\subsection{Experimental Settings}
\label{subsec:settings}

We use the ENCOW14A corpus~\cite{Schaefer:12,Schaefer:15} with
approx. 14.5 billion tokens for training the hierarchical embeddings
and the default SGNS model.  We train our model with 100 dimensions, a window size of 5,
15 negative samples, and 0.025 as the learning rate. The threshold
$\theta$ is set to 0.05.
The hypernymy resource for nouns comprises $105,020$ hyponyms, $24,925$
hypernyms, and $1,878,484$ hyponym--hypernym pairs. The hypernymy
resource for verbs consists of $11,328$ hyponyms, $4,848$ hypernyms,
and $130,350$ hyponym--hypernym pairs.
\begin{table}[]
\centering
\resizebox{0.99\linewidth}{!}{
\begin{tabular}{|l|l|r|c|}
\hline
\multicolumn{1}{|c}{\textbf{Dataset}} & \multicolumn{1}{|c}{\textbf{Relation}} & \multicolumn{1}{|c|}{\textbf{\#Instance}} & \textbf{Total}          \\  \hline
\multirow{8}{*}{BLESS}                 & hypernymy                               & 1,337                                    & \multirow{8}{*}{26,554} \\
                                       & meronymy                                & 2,943                                    &                         \\
                                       & coordination                           & 3,565                                    &                         \\
                                       & event                                  & 3,824                                    &                         \\
                                       & attribute                              & 2,731                                    &                         \\
                                       & random-n                               & 6,702                                    &                         \\
                                       & random-j                               & 2,187                                    &                         \\
                                       & random-v                               & 3,265                                    &                         \\ \hline
\multirow{5}{*}{EVALution}             & hypernymy                               & 3,637                                    & \multirow{5}{*}{13,465} \\
                                       & meronymy                                & 1,819                                    &                         \\
                                       & attribute                              & 2,965                                    &                         \\
                                       & synonymy                                & 1,888                                    &                         \\
                                       & antonymy                                & 3,156                                    &                         \\ \hline
\multirow{3}{*}{Lenci\&Benotto}          & hypernymy                               & 1,933                                    & \multirow{3}{*}{5,010}  \\
                                       & synonymy                                & 1,311                                    &                         \\
                                       & antonymy                                & 1,766                                    &                         \\ \hline
\multirow{2}{*}{Weeds}                 & hypernymy                               & 1,469                                    & \multirow{2}{*}{2,928}  \\
                                       & coordination                           & 1,459                                    &                         \\ \hline
\end{tabular}
}
\caption{Details of the semantic relations and the number of instances in each dataset.}
\label{tbl:ap-datasets}
\vspace{+0.2cm}
\end{table}

\subsection{Unsupervised Hypernymy Detection and Directionality}
\label{subsec:unsupervised-detection-directionality}
\begin{figure*}[t]
  \centering
  \begin{subfigure}{.50\textwidth}
    \centering
    \includegraphics[trim=0 0.5cm 0cm 0cm 0,width=0.99\linewidth]{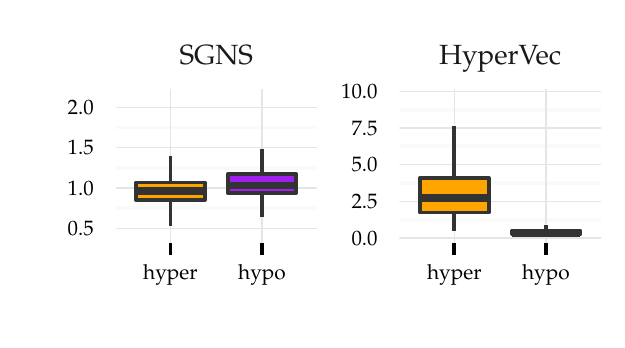}
    \caption{Directionality task: hypernym vs. hyponym. }
    \label{fig:sub1_direc}
  \end{subfigure}%
  \begin{subfigure}{.50\textwidth}
    \centering	\includegraphics[trim=0 0.5cm 0cm 0cm 0,width=0.99\linewidth]{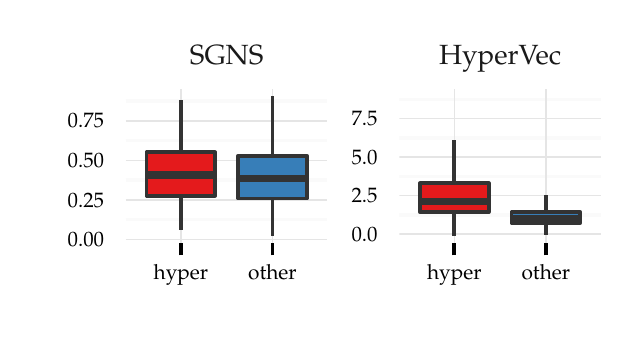}
    \caption{Hypernymy detection: hypernymy vs. other relations. 
    }	
    \label{fig:sub2_other}
  \end{subfigure}
  \caption{Comparing $\mathit{SGNS}$ and $\mathit{HyperVec}$ on binary classification tasks. The y-axis shows the magnitude values of the vectors.}
  \label{fig:kiela_exp}
\end{figure*}

In this section, we assess our model on two experimental setups: i) a
ranking retrieval setup that expects hypernymy pairs to have a higher
similarity score than instances from other semantic relations; ii) a
classification setup that requires both hypernymy detection and
directionality.

\subsubsection{Ranking Retrieval}
\label{subsubsec:unsupervised-ranking} 

\newcite{Shwartz:17} conducted an extensive evaluation of a large
number of unsupervised distributional measures for hypernymy ranking
retrieval proposed in previous
work~\cite{Weeds/Weir:03,Santus:14,Clarke:09,Kotlerman:10,Lenci/Benotto:12,Santus:16}. The
evaluation was performed on four semantic relation datasets: \textbf{\textsc{BLESS}}~\cite{Baroni/Lenci:11}, \textbf{\textsc{Weeds}}~\cite{Weeds:04}, 
\textbf{\textsc{EVALution}}~\cite{Santus:15}, and \textbf{\textsc{Lenci\&Benotto}}~\cite{Benotto:15}.
Table~\ref{tbl:ap-datasets} describes the detail of these datasets
in terms of the semantic relations and the number of instances.
The Average Precision (AP) ranking measure is used to evaluate the
performance of the measures.
\begin{table}[]
  \centering
  \resizebox{0.99\linewidth}{!}{
    \begin{tabular}{|c|l|cc|}
      \hline
      \textbf{Dataset}              & \multicolumn{1}{c|}{\textbf{Hypernymy vs.}} & \textbf{Baseline} & \textbf{HyperScore} \\ \hline
      \multirow{5}{*}{EVALution}    & other relations                            & 0.353             & \textbf{0.538}      \\
                                    & meronymy                                    & 0.675             & \textbf{0.811}      \\
                                    & attribute                                  & 0.651             & \textbf{0.800}      \\
                                    & antonymy                                    & 0.55              & \textbf{0.743}      \\
                                    & synonymy                                    & 0.657             & \textbf{0.793}      \\ \hline
      \multirow{5}{*}{BLESS}        & other relations                            & 0.051             & \textbf{0.454}      \\
                                    & meronymy                                    & 0.76              & \textbf{0.913}      \\
                                    & coordination                                & 0.537             & \textbf{0.888}      \\
                                    & attribute                                  & 0.74              & \textbf{0.918}      \\
                                    & event                                      & \textbf{0.779}    & 0.620               \\ \hline
      \multirow{3}{*}{Lenci\&Benotto} & other relations                            & 0.382             & \textbf{0.574}      \\
                                    & antonymy                                    & 0.624             & \textbf{0.696}      \\
                                    & synonymy                                    & 0.725             & \textbf{0.751}      \\ \hline
      Weeds                         & coordination                                      & 0.441             & \textbf{0.850}      \\ \hline
    \end{tabular}
  }
\caption{AP results of \textit{HyperScore} in comparison to state-of-the-art measures.}
\label{tbl:unsupervised-measures}
\vspace{-0.4cm}
\end{table}

In comparison to the state-of-the-art unsupervised measures compared by
\newcite{Shwartz:17} (henceforth, baseline models), we apply our
unsupervised measure $\mathit{HyperScore}$ (Equation~\ref{eq7}) to
rank hypernymy against other
relations. Table~\ref{tbl:unsupervised-measures} presents the results
of using $\mathit{HyperScore}$ vs. the best baseline models, across
datasets. When detecting hypernymy among all other relations (which is
the most challenging task), $\mathit{HyperScore}$ significantly
outperforms all baseline variants on all datasets. The strongest
difference is reached on the BLESS dataset, where
$\mathit{HyperScore}$ achieves an improvement of 40$\%$ AP score over
the best baseline model. When ranking hypernymy in comparison to a
single other relation, $\mathit{HyperScore}$ also improves over the
baseline models, except for the \textit{event} relation in the BLESS
dataset. We assume that this is due to the different parts-of-speech
(adjective and noun) involved in the relation, where
$\mathit{HyperVec}$ fails to establish a hierarchy.

\subsubsection{Classification}
\label{subsubsec:unsupervised-classification}

In this setup, we rely on three datasets of semantic relations, which
were all used in various state-of-the-art approaches before, and
brought together for hypernymy evaluation by
\newcite{Kiela:15}. \textbf{(i)} A subset of \textbf{\textsc{BLESS}}
contains 1,337 hyponym-hypernym pairs. The task is to predict the
directionality of hypernymy within a binary classification.
Our approach requires no threshold; we only need to compare the
magnitudes of the two words and to assign the hypernym label to the
word with the larger magnitude.
Figure~\ref{fig:sub1_direc} indicates that the magnitude values of the
$\mathit{SGNS}$ model cannot distinguish between a hyponym and a
hypernym, while the hierarchical embeddings provide a larger magnitude
for the hypernym. \textbf{(ii)} Following \newcite{Weeds:14}, we
conduct a binary classification with a subset of 1,168 \textsc{BLESS}
word pairs. In this dataset (\textbf{\textsc{WBLESS}}), one class is
represented by hyponym--hypernym pairs, and the other class is a
combination of reversed hypernym--hyponym pairs, plus additional
holonym-meronym pairs, co-hyponyms and randomly matched nouns. For
this classification we make use of our $\mathit{HyperScore}$ measure
that ranks hypernymy pairs higher than other relation pairs. A
threshold decides about the splitting point between the two classes:
\textit{hyper} vs.  \textit{other}. Instead of using a manually
defined threshold as done by \newcite{Kiela:15}, we decided to run
$1\,000$ iterations which randomly sampled only 2\% of the available
pairs for learning a threshold, using the remaining 98\% for test
purposes. We present average accuracy results across all iterations.
Figure~\ref{fig:sub2_other} compares the default cosine similarities
between the relation pairs (as applied by $\mathit{SGNS}$) and
$\mathit{HyperScore}$ (as applied by $\mathit{HyperVec}$) on this
task. Using $\mathit{HyperScore}$, the class ``hyper'' can clearly be
distinguished from the class ``other''. \textbf{(iii)
  \textbf{\textsc{BIBLESS}}} represents the most challenging dataset; the
relation pairs from \textsc{WBLESS} are split into three classes
instead of two: hypernymy pairs, reversed hypernymy pairs, and other
relation pairs. In this case, we perform a three-way
classification. We apply the same technique as used for the
\textsc{WBLESS} classification, but in cases where we classify
\textit{hyper} we additionally classify the hypernymy direction, to
decide between hyponym--hypernym pairs and reversed hypernym--hyponym
pairs.
\begin{table}[]	
	\centering
	\resizebox{\linewidth}{!}{ 
		\begin{tabular}{lrrr}
			\toprule \toprule
			& \textsc{BLESS} &  \textsc{WBLESS} &  \textsc{BIBLESS} \\ \midrule  
			\newcite{Kiela:15} & 0.88 & 0.75 & {0.57} \\ 
			\newcite{Santus:14} & 0.87 & ----- & ----- \\ 
			\newcite{Weeds:14} &  -----& 0.75 & ----- \\  \midrule
			$\mathit{SGNS}$  & 0.44 & 0.48 & 0.34 \\ 
			$\mathit{HyperVec}$  & \textbf{0.92} & \textbf{0.87} & {\textbf{0.81}} \\ \bottomrule \bottomrule
	\end{tabular} }
	\caption{Accuracy for hypernymy directionality.}
	\label{results_exp_direction} 
\end{table} 

Table~\ref{results_exp_direction} compares our results against related
work. $\mathit{HyperVec}$ outperforms all other methods on all three
tasks. In addition we see again that an unmodified $\mathit{SGNS}$
model cannot solve any of the three tasks.


\subsection{Supervised Hypernymy Detection}
\label{subsec:supervised-hypernymy-detection}

For supervised hypernymy detection, we make use of the two datasets:
the full \textbf{\textsc{BLESS}} dataset, and
\textbf{\textsc{ENTAILMENT}}~\cite{Baroni:12}, containing 2,770
relation pairs in total, including 1,385 hypernym pairs and 1,385
other relations pairs.
We follow the same procedure as~\newcite{Yu:15} and \newcite{Tuan:16}
to assess $\mathit{HyperVec}$ on the two datasets. Regarding BLESS, we
extract pairs for four types of relations: hypernymy, meronymy,
co-hyponymy (or \textit{coordination}), and add the random relation
for nouns. For the evaluation, we randomly select one concept and its
relatum for testing, and train the supervised model on the 199
remaining concepts and its relatum. We then report the average
accuracy across all concepts. For the ENTAILMENT dataset, we randomly
select one hypernym pair for testing and train on all remaining
hypernym pairs. Again, we report the average accuracy across all
hypernyms.

We apply an SVM classifier to detect hypernymy based on
$\mathit{HyperVec}$. Given a hyponym--hypernym pair ($u,v$), we
concatenate four components to construct the vector for a pair ($u,v$)
as follows: the vector difference between hypernym and hyponym
($\vec{v} - \vec{u}$); the cosine similarity between the hypernym and
hyponym vectors ($cos(\vec{u},\vec{v})$); the magnitude of the hyponym
($\|\vec{u}\|$); and the magnitude of the hypernym
($\|\vec{v}\|$). The resulting vector is fed into the SVM classifier
to detect hypernymy. Similar to the two previous works, we train the
SVM classifier with the RBF kernel, $\lambda=0.03125$, and the penalty
$C=8.0$.

Table~\ref{tbl:supervised-detection} shows the performance of
$\mathit{HyperVec}$ and the two baseline models reported
by~\newcite{Tuan:16}. $\mathit{HyperVec}$ slightly outperforms the
method of~\newcite{Tuan:16} on the BLESS dataset, and is equivalent to
the performance of their method on the ENTAILMENT dataset. In
comparison to the method of~\newcite{Yu:15}, $\mathit{HyperVec}$
achieves significant improvements.
\begin{table}[]
  \centering
  \resizebox{0.99\linewidth}{!}{
    \begin{tabular}{lcc}
      \toprule\toprule
      \textbf{Models} & \textbf{BLESS} & \textbf{ENTAILMENT} \\ \midrule
      \newcite{Yu:15}          & 0.90           & 0.87                \\
      \newcite{Tuan:16}        & 0.93           & 0.91                \\ \midrule
      $\mathit{HyperVec}$                      & \textbf{0.94}  & 0.91                    \\ \toprule
    \end{tabular}
  }
  \caption{Classification results for BLESS and ENTAILMENT in terms of accuracy.}
  \label{tbl:supervised-detection}
  \vspace{-0.4cm}
\end{table} 

\subsection{Graded Lexical Entailment}
\label{subsec:graded-lexical-entailment}

In this experiment, we apply $\mathit{HyperVec}$ to the dataset of
graded lexical entailment, $\textit{HyperLex}$, as introduced by
\newcite{Vulic:16}. The $\textit{HyperLex}$ dataset provides soft lexical entailment on a continuous scale, rather than
simplifying into a binary decision.
$\textit{HyperLex}$ contains 2,616 word pairs across seven semantic
relations and two word classes (nouns and verbs). Each word pair is
rated by a score that indicates the strength of the semantic relation
between the two words. For example, the score of the hypernym pair
\textit{duck--animal} is 5.9 out of 6.0, while the score of the
reversed pair \textit{animal--duck} is only 1.0.

We compared $\mathit{HyperScore}$ against the most prominent
state-of-the-art hypernymy and lexical entailment models from previous
work: 
\begin{itemize}
	\itemsep0em 
	\item Directional entailment measures
	(DEM)~\cite{Weeds/Weir:03,Weeds:04,Clarke:09,Kotlerman:10,Lenci/Benotto:12}
	\item Generality measures (SQLS)~\cite{Santus:14}
	\item Visual generality measures (VIS)~\cite{Kiela:15}
	\item Consideration of concept frequency ratio (FR) \cite{Vulic:16}
	\item WordNet-based similarity measures 
	(WN)~\cite{Wu/Palmer:94,Pedersen:04}
	\item Order embeddings 
	(OrderEmb)~\cite{Vendrov:16}
	\item Skip-gram embeddings 
	(SGNS)~\cite{Mikolov:13,Levy/Goldberg:14}
	\item Embeddings fine-tuned to a
	paraphrase database with linguistic constraints (PARAGRAM)~\cite{Mrksic:16}
	\item Gaussian embeddings (Word2Gauss)~\cite{Vilnis/McCallum:15}
\end{itemize}
The performance of the models
is assessed through Spearman's rank-order correlation coefficient
$\rho$~\cite{Siegel/Castellan:88}, comparing the ranks of the models' scores and the human
judgments for the given word pairs.
\begin{table}[]
  \centering
    \begin{tabular}{lclc}
      \toprule \toprule
      \multicolumn{2}{c}{\textbf{Measures}}                     & \multicolumn{2}{c}{\textbf{Embeddings}}              \\ \midrule
      \multicolumn{1}{c}{\textbf{Model}} & $\rho$ & \multicolumn{1}{c}{\textbf{Model}} & $\rho$ \\ \midrule
      FR                                   & 0.279                & SGNS                                & 0.205                \\
      DEM                                  & 0.180                & PARAGRAM                            & 0.320                \\
      SLQS                                 & 0.228                & OrderEmb                            & 0.191                \\
      WN                                   & 0.234                & Word2Gauss                          & 0.206                \\
      VIS                                  & 0.209                & $\mathit{HyperScore}$               & \textbf{0.540}        \\ \bottomrule \bottomrule
    \end{tabular}
  \caption{Results ($\rho$) of $\mathit{HyperScore}$ and state-of-the-art measures and word embedding models on graded lexical entailment.}
\label{tbl:graded-hypernymy}
\vspace{-0.4cm}
\end{table}

Table~\ref{tbl:graded-hypernymy} shows that $\mathit{HyperScore}$
significantly outperforms both state-of-the-art measures and word
embedding models. $\mathit{HyperScore}$ outperforms even the
previously best word embedding model PARAGRAM by .22, and the
previously best measures FR by .27. The reason that
$\mathit{HyperVec}$ outperforms all other models is that the hierarchy
between hypernym and hypornym within $\mathit{HyperVec}$
differentiates hyponym--hypernym pairs from hypernym--hyponym
pairs. For example, the $\mathit{HyperScore}$ for the pairs
\textit{duck--animal} and \textit{animal--duck} are 3.02 and 0.30,
respectively. Thus, the magnitude proportion of the hypernym--hyponym
pair \textit{duck--animal} is larger than that for the pair
\textit{animal--duck}.

\section{Generalizing Hypernymy}


Having demonstrated the general abilities of $\mathit{HyperVec}$, this
final section explores its potential for generalization in two
different ways, (i) by relying on a small seed set only, rather than
using a large set of training data; and (ii) by projecting
$\mathit{HyperVec}$ to other languages.

\paragraph{Hypernymy Seed Generalization:}

We utilize only a small hypernym set from the hypernymy resource to
train $\mathit{HyperVec}$, relying on 200 concepts from the
\textsc{BLESS} dataset. The motivation behind using these concepts is threefold: i) these concepts are distinct and unambiguous noun concepts; ii) the concepts were equally divided between living and non-living entities; iii) concepts have been grouped into 17 broader classes. 
Based on the seed set, we collected the hyponyms of each
concept from WordNet, and then re-trained $\mathit{HyperVec}$. On the
hypernymy ranking retrieval task
(Section~\ref{subsubsec:unsupervised-ranking}), $\mathit{HyperScore}$
outperforms the baselines across all datasets (cf. Table~1) with AP
values of 0.39, 0.448, and 0.585 for EVALution, LenciBenotto, and
Weeds, respectively. For the graded lexical entailment task
(Section~\ref{subsec:graded-lexical-entailment}),
$\mathit{HyperScore}$ obtains a correlation of $\rho=0.30$,
outperforming all models except for PARAGRAM with
$\rho=0.32$. Overall, the results show that $\mathit{HyperVec}$ is
indeed able to generalize hypernymy from small seeds of training data.

\paragraph{Generalizing Hypernymy across Languages:}

We assume that hypernymy detection can be improved across languages by
projecting representations from any arbitrary language into our
modified English $\mathit{HyperVec}$ space.  We conduct experiments
for German and Italian, where the language-specific representations
are obtained using the same hyper-parameter settings as for our
English $\mathit{SGNS}$ model (cf. Section~\ref{subsec:settings}). As
corpus resource we relied on Wikipedia dumps\footnote{The Wikipedia
  dump for German and Italian were both downloaded in January
  2017.}. Note that we do not use any additional resource, such as the
German or Italian WordNet, to tune the embeddings for hypernymy
detection.  Based on the representations, a mapping function between a
source language (German, Italian) and our English $\mathit{HyperVec}$
space is learned, by relying on the least-squares error method from
previous work using cross-lingual data \cite{MikolovLS13} and
different modalities \cite{LazaridouDB15}.

To learn a mapping function between two languages, a one-to-one
correspondence (word translations) between two sets of vectors is
required. We obtained these translations by using the parallel
Europarl\footnote{\scriptsize \url{http://www.statmt.org/europarl/}} V7 corpus for
German--English and Italian--English. Word alignment counts were
extracted using $\mathit{fast\_align}$ \cite{Dyer2013ASF}. We then
assigned each source word to the English word with the maximum number
of alignments in the parallel corpus. We could match 25,547 pairs for
DE$\to$EN and 47,475 pairs for IT$\to$EN.

Taking the aligned subset of both spaces, we assume that $X$ is the
matrix obtained by concatenating all source vectors, and likewise $Y$
is the matrix obtained by concatenating all corresponding English
elements. Applying the $\ell2$-regularized least-squares error
objective can be described using the following equation:
\begin{equation}
 \hat{\textbf{W}}  = \underset{\textbf{W} \in \mathbb{R}^{d1 \times d2}}{\mathrm{argmin}}   \|\textbf{XW}-\textbf{Y}\| + \lambda \| \textbf{W}\|
\end{equation} 
%
Although we learn the mapping only on a subset of aligned words, it
allows us to project every word in a source vocabulary to its English
$\mathit{HyperVec}$ position by using $\textbf{W}$.

Finally we compare the original representations and the mapped
representation on the hypernymy ranking retrieval task (similar to
Section~\ref{subsubsec:unsupervised-ranking}). As gold resources we
relied on German and Italian nouns pairs. For German we used the 282
German pairs collected via Amazon Mechanical Turk by
\newcite{Scheible/Schulteimwalde:2014}. The 1,350 Italian pairs
were collected via Crowdflower by \newcite{Sucameli15} in the same
way. Both collections contain hypernymy, antonymy and synonymy
pairs. As before, we evaluate the ranking by AP, and we compare the
cosine of the unmodified default representations against the
$\mathit{HyperScore}$ of the projected representations.

\begin{table}[htbp]	
  \centering
  \resizebox{\linewidth}{!}{ 
    \begin{tabular}{lccc}
      \toprule \toprule
      German   &  Hyp/All  & Hyp/Syn & Hyp/Ant \\
      DE-$\mathit{SGNS}$ & 0.28 & 0.48 & 0.40 \\ 
      DE$\to$EN$\mathit{HyperVec}$ & \textbf{0.37} & \textbf{0.65} & \textbf{0.47} \\ 
      \midrule
      Italian  &  & &  \\ 
      IT-$\mathit{SGNS}$ & 0.38 & 0.50 & 0.60 \\ 
      IT$\to$EN$\mathit{HyperVec}$ & \textbf{0.44} & \textbf{0.57} & \textbf{0.65} \\ \bottomrule \bottomrule
    \end{tabular} }
  \caption{AP results across languages, comparing $\mathit{SGNS}$ and the projected representations.}
  \label{results_exp_mapping}
  \vspace{-0.2cm}
\end{table}
	
The results are shown in Table~\ref{results_exp_mapping}. We clearly
see that for both languages the default $\mathit{SGNS}$ embeddings do
not provide higher similarity scores for hypernymy pairs (except for
Italian Hyp/Ant), but both languages provide higher scores when we map
the embeddings into the English $\mathit{HyperVec}$ space.


\section{Conclusion}

This paper proposed a novel neural model $\mathit{HyperVec}$ to learn
hierarchical embeddings for hypernymy. $\mathit{HyperVec}$ has been
shown to strengthen hypernymy similarity, and to capture the
distributional hierarchy of hypernymy. Together with a newly proposed
unsupervised measure $\mathit{HyperScore}$ our experiments
demonstrated (i) significant improvements against state-of-the-art
measures, and (ii) the capability to generalize hypernymy and learn
the relation instead of memorizing \textit{prototypical hypernyms}.

\section*{Acknowledgments}
The research was supported by the Ministry of Education and Training of the Socialist Republic of Vietnam (Scholarship 977/QD-BGDDT; Kim-Anh Nguyen), the DFG Collaborative Research Centre SFB 732 (Kim-Anh Nguyen, Maximilian K\"oper, Ngoc Thang Vu), and the DFG Heisenberg Fellowship SCHU-2580/1 (Sabine Schulte im Walde). We would like to thank three anonymous reviewers for their comments and suggestions. 

\bibliography{emnlp2017}

\begin{thebibliography}{51}
\expandafter\ifx\csname natexlab\endcsname\relax\def\natexlab#1{#1}\fi

\bibitem[{Baroni et~al.(2012)Baroni, Bernardi, Do, and Shan}]{Baroni:12}
Marco Baroni, Raffaella Bernardi, Ngoc-Quynh Do, and Chung-chieh Shan. 2012.
\newblock Entailment above the word level in distributional semantics.
\newblock In \emph{Proceedings of the 13th Conference of the European Chapter
  of the Association for Computational Linguistics (EACL)}, pages 23--32,
  Avignon, France.

\bibitem[{Baroni and Lenci(2011)}]{Baroni/Lenci:11}
Marco Baroni and Alessandro Lenci. 2011.
\newblock How we blessed distributional semantic evaluation.
\newblock In \emph{Proceedings of the GEMS 2011 Workshop on GEometrical Models
  of Natural Language Semantics (GEMS)}, pages 1--10, Edinburgh, Scotland.

\bibitem[{Benotto(2015)}]{Benotto:15}
Giulia Benotto. 2015.
\newblock \emph{Distributional models for semantic relations: A study on
  hyponymy and antonymy}.
\newblock Ph.D. thesis, University of Pisa.

\bibitem[{Biran and McKeown(2013)}]{Biran/McKeown:13}
Or~Biran and Kathleen McKeown. 2013.
\newblock Classifying taxonomic relations between pairs of wikipedia articles.
\newblock In \emph{Proceddings of Sixth International Joint Conference on
  Natural Language Processing (IJCNLP)}, pages 788--794, Nagoya, Japan.

\bibitem[{Clarke(2009)}]{Clarke:09}
Daoud Clarke. 2009.
\newblock Context-theoretic semantics for natural language: An overview.
\newblock In \emph{Proceedings of the Workshop on Geometrical Models of Natural
  Language Semantics (GEMS)}, pages 112--119, Athens, Greece.

\bibitem[{Dagan et~al.(2013)Dagan, Roth, Sammons, and Zanzotto}]{Dagan:13}
Ido Dagan, Dan Roth, Mark Sammons, and Fabio~Massimo Zanzotto. 2013.
\newblock \emph{Recognizing Textual Entailment: Models and Applications}.
\newblock Synthesis Lectures on Human Language Technologies.

\bibitem[{Dyer et~al.(2013)Dyer, Chahuneau, and Smith}]{Dyer2013ASF}
Chris Dyer, Victor Chahuneau, and Noah~A. Smith. 2013.
\newblock {A} {S}imple, {F}ast, and {E}ffective {R}eparameterization of {IBM}
  {M}odel 2.
\newblock In \emph{Proceedings of the North American Chapter of the Association
  for Computational Linguistics: Human Language Technologies (NAACL)}, pages
  644--648, Atlanta, USA.

\bibitem[{Fallucchi and Zanzotto(2011)}]{Fallucchi/Zanzotto:2011}
Francesca Fallucchi and Fabio~Massimo Zanzotto. 2011.
\newblock Inductive probabilistic taxonomy learning using singular value
  decomposition.
\newblock \emph{Natural Language Engineering}, 17(1):71--94.

\bibitem[{Fellbaum(1998)}]{Fellbaum:98}
Christiane Fellbaum, editor. 1998.
\newblock \emph{WordNet -- An Electronic Lexical Database}.
\newblock Language, Speech, and Communication. MIT Press, Cambridge, MA.

\bibitem[{Firth(1957)}]{Firth:57}
John~R. Firth. 1957.
\newblock \emph{Papers in Linguistics 1934-51}.
\newblock Longmans, London, UK.

\bibitem[{Geffet and Dagan(2005)}]{Geffet/Dagan:05}
Maayan Geffet and Ido Dagan. 2005.
\newblock The distributional inclusion hypotheses and lexical entailment.
\newblock In \emph{Proceedings of the 43rd Annual Meeting of the Association
  for Computational Linguistics (ACL)}, pages 107--114, Michigan, US.

\bibitem[{Harris(1954)}]{Harris:54}
Zellig~S. Harris. 1954.
\newblock Distributional structure.
\newblock \emph{Word}, 10(23):146--162.

\bibitem[{Kiela et~al.(2015)Kiela, Rimell, Vuli\'{c}, and Clark}]{Kiela:15}
Douwe Kiela, Laura Rimell, Ivan Vuli\'{c}, and Stephen Clark. 2015.
\newblock Exploiting image generality for lexical entailment detection.
\newblock In \emph{Proceedings of the 53rd Annual Meeting of the Association
  for Computational Linguistics and the 7th International Joint Conference on
  Natural Language Processing (ACL)}, pages 119--124, Beijing, China.

\bibitem[{Kotlerman et~al.(2010)Kotlerman, Dagan, Szpektor, and
  Zhitomirsky{-}Geffet}]{Kotlerman:10}
Lili Kotlerman, Ido Dagan, Idan Szpektor, and Maayan Zhitomirsky{-}Geffet.
  2010.
\newblock Directional distributional similarity for lexical inference.
\newblock \emph{Natural Language Engineering}, 16(4):359--389.

\bibitem[{Lazaridou et~al.(2015)Lazaridou, Dinu, and Baroni}]{LazaridouDB15}
Angeliki Lazaridou, Georgiana Dinu, and Marco Baroni. 2015.
\newblock {H}ubness and {P}ollution: {D}elving into {C}ross-{S}pace {M}apping
  for {Z}ero-{S}hot {L}earning.
\newblock In \emph{Proceedings of the 53rd Annual Meeting of the Association
  for Computational Linguistics (ACL)}, pages 270--280, Beijing, China.

\bibitem[{Lenci and Benotto(2012)}]{Lenci/Benotto:12}
Alessandro Lenci and Giulia Benotto. 2012.
\newblock Identifying hypernyms in distributional semantic spaces.
\newblock In \emph{{*SEM 2012}: The First Joint Conference on Lexical and
  Computational Semantics -- Volume 1: Proceedings of the main conference and
  the shared task, and Volume 2: Proceedings of the Sixth International
  Workshop on Semantic Evaluation (SemEval)}, pages 75--79, Montr\'{e}al,
  Canada.

\bibitem[{Levy and Goldberg(2014)}]{Levy/Goldberg:14}
Omer Levy and Yoav Goldberg. 2014.
\newblock Neural word embedding as implicit matrix factorization.
\newblock In \emph{Proceddings of the 27th International Conference on Advances
  in Neural Information Processing Systems (NIPS)}, pages 2177--2185,
  Montr\'{e}al, Canada.

\bibitem[{Levy et~al.(2015)Levy, Remus, Biemann, and Dagan}]{Levy:15}
Omer Levy, Steffen Remus, Chris Biemann, and Ido Dagan. 2015.
\newblock Do supervised distributional methods really learn lexical inference
  relations?
\newblock In \emph{Proceedings of the 2015 Conference of the North American
  Chapter of the Association for Computational Linguistics: Human Language
  Technologies (NAACL)}, pages 970--976, Denver, Colorado.

\bibitem[{Mikolov et~al.(2013{\natexlab{a}})Mikolov, Le, and
  Sutskever}]{MikolovLS13}
Tomas Mikolov, Quoc~V. Le, and Ilya Sutskever. 2013{\natexlab{a}}.
\newblock {E}xploiting {S}imilarities among {L}anguages for {M}achine
  {T}ranslation.
\newblock \emph{CoRR}, abs/1309.4168.

\bibitem[{Mikolov et~al.(2013{\natexlab{b}})Mikolov, Sutskever, Chen, Corrado,
  and Dean}]{Mikolov:13}
Tomas Mikolov, Ilya Sutskever, Kai Chen, Greg Corrado, and Jeffrey Dean.
  2013{\natexlab{b}}.
\newblock Distributed representations of words and phrases and their
  compositionality.
\newblock In \emph{Proceedings of the 26th International Conference on Advances
  in Neural Information Processing Systems (NIPS)}, pages 3111--3119, Lake
  Tahoe, Nevada, US.

\bibitem[{Miller(1995)}]{Miller:95}
George~A. Miller. 1995.
\newblock {WordNet}: A lexical database for {English}.
\newblock \emph{Communications of the ACM}, 38(11):39--41.

\bibitem[{Miller and Fellbaum(1991)}]{Miller/Fellbaum:91}
George~A. Miller and Christiane Fellbaum. 1991.
\newblock Semantic networks of english.
\newblock \emph{Cognition}, 41:197--229.

\bibitem[{Mrk{\v{s}}i{\'{c}} et~al.(2016)Mrk{\v{s}}i{\'{c}},
  {\'O}~S{\'e}aghdha, Thomson, Ga{\v{s}}i{\'{c}}, Rojas-Barahona, Su, Vandyke,
  Wen, and Young}]{Mrksic:16}
Nikola Mrk{\v{s}}i{\'{c}}, Diarmuid {\'O}~S{\'e}aghdha, Blaise Thomson, Milica
  Ga{\v{s}}i{\'{c}}, M.~Lina Rojas-Barahona, Pei-Hao Su, David Vandyke,
  Tsung-Hsien Wen, and Steve Young. 2016.
\newblock Counter-fitting word vectors to linguistic constraints.
\newblock In \emph{Proceedings of the 2016 Conference of the North American
  Chapter of the Association for Computational Linguistics: Human Language
  Technologies (NAACL-HLT)}, pages 142--148, San Diego, California.

\bibitem[{Murphy(2002)}]{Murphy:02}
Gregory Murphy. 2002.
\newblock \emph{The Big Book of Concepts}.
\newblock MIT Press, Cambridge, MA, USA.

\bibitem[{Navigli et~al.(2011)Navigli, Velardi, and Faralli}]{Navigli:11}
Roberto Navigli, Paola Velardi, and Stefano Faralli. 2011.
\newblock A graph-based algorithm for inducing lexical taxonomies from scratch.
\newblock In \emph{Proceedings of the Twenty-Second International Joint
  Conference on Artificial Intelligence (IJCAI)}, pages 1872--1877, Barcelona,
  Catalonia, Spain.

\bibitem[{Nguyen et~al.(2016)Nguyen, Schulte~im Walde, and Vu}]{Nguyen:16}
Kim~Anh Nguyen, Sabine Schulte~im Walde, and Ngoc~Thang Vu. 2016.
\newblock Integrating distributional lexical contrast into word embeddings for
  antonym-synonym distinction.
\newblock In \emph{Proceedings of the 54th Annual Meeting of the Association
  for Computational Linguistics (ACL)}, pages 454–--459, Berlin, Germany.

\bibitem[{Pedersen et~al.(2004)Pedersen, Patwardhan, and
  Michelizzi}]{Pedersen:04}
Ted Pedersen, Siddharth Patwardhan, and Jason Michelizzi. 2004.
\newblock Wordnet: : Similarity - measuring the relatedness of concepts.
\newblock In \emph{Proceedings of the 19th National Conference on Artificial
  Intelligence, Sixteenth Conference on Innovative Applications of Artificial
  Intelligence (AAAI)}, pages 1024--1025, California, USA.

\bibitem[{Pennington et~al.(2014)Pennington, Socher, and
  Manning}]{Pennington:14}
Jeffrey Pennington, Richard Socher, and Christopher~D. Manning. 2014.
\newblock Glove: Global vectors for word representation.
\newblock In \emph{Proceedings of the 2014 Conference on Empirical Methods in
  Natural Language Processing (EMNLP)}, pages 1532--1543, Doha, Qatar.

\bibitem[{Rimell(2014)}]{Rimell:14}
Laura Rimell. 2014.
\newblock Distributional lexical entailment by topic coherence.
\newblock In \emph{Proceedings of the 14th Conference of the European Chapter
  of the Association for Computational Linguistics (EACL)}, pages 511--519,
  Gothenburg, Sweden.

\bibitem[{Roller et~al.(2014)Roller, Erk, and Boleda}]{Roller:14}
Stephen Roller, Katrin Erk, and Gemma Boleda. 2014.
\newblock {Inclusive yet selective: Supervised distributional hypernymy
  detection}.
\newblock In \emph{Proceedings of the 25th International Conference on
  Computational Linguistics (COLING)}, pages 1025--1036, Dublin, Ireland.

\bibitem[{Santus et~al.(2016)Santus, Lenci, Chiu, Lu, and Huang}]{Santus:16}
Enrico Santus, Alessandro Lenci, Tin{-}Shing Chiu, Qin Lu, and Chu{-}Ren Huang.
  2016.
\newblock Unsupervised measure of word similarity: How to outperform
  co-occurrence and vector cosine in vsms.
\newblock In \emph{Proceedings of the Thirtieth Conference on Artificial
  Intelligence AAAI)}, pages 4260--4261, Arizona, USA.

\bibitem[{Santus et~al.(2014)Santus, Lenci, Lu, and Walde}]{Santus:14}
Enrico Santus, Alessandro Lenci, Qin Lu, and Sabine Schulte~Im Walde. 2014.
\newblock Chasing hypernyms in vector spaces with entropy.
\newblock In \emph{Proceedings of the 14th Conference of the European Chapter
  of the Association for Computational Linguistics (EACL)}, pages 38--42,
  Gothenburg, Sweden.

\bibitem[{Santus et~al.(2015)Santus, Yung, Lenci, and Huang}]{Santus:15}
Enrico Santus, Frances Yung, Alessandro Lenci, and Chu-Ren Huang. 2015.
\newblock Evalution 1.0: an evolving semantic dataset for training and
  evaluation of distributional semantic models.
\newblock In \emph{Proceedings of the 4th Workshop on Linked Data in
  Linguistics: Resources and Applications}, Beijing, China.

\bibitem[{Sch\"afer(2015)}]{Schaefer:15}
Roland Sch\"afer. 2015.
\newblock Processing and querying large web corpora with the {COW14}
  architecture.
\newblock In \emph{Proceedings of the 3rd Workshop on Challenges in the
  Management of Large Corpora}, pages 28--34, Lancaster, UK.

\bibitem[{Sch\"afer and Bildhauer(2012)}]{Schaefer:12}
Roland Sch\"afer and Felix Bildhauer. 2012.
\newblock Building large corpora from the web using a new efficient tool chain.
\newblock In \emph{Proceedings of the 8th International Conference on Language
  Resources and Evaluation}, pages 486--493, Istanbul, Turkey.

\bibitem[{Scheible and Schulte~im Walde(2014)}]{Scheible/Schulteimwalde:2014}
Silke Scheible and Sabine Schulte~im Walde. 2014.
\newblock {A} {D}atabase of {P}aradigmatic {S}emantic {R}elation {P}airs for
  {G}erman {N}ouns, {V}erbs, and {A}djectives.
\newblock In \emph{Proceedings of Workshop on Lexical and Grammatical Resources
  for Language Processing}, pages 111--119, Dublin, Ireland.

\bibitem[{Shwartz et~al.(2017)Shwartz, Santus, and Schlechtweg}]{Shwartz:17}
Vered Shwartz, Enrico Santus, and Dominik Schlechtweg. 2017.
\newblock Hypernyms under siege: Linguistically-motivated artillery for
  hypernymy detection.
\newblock In \emph{Proceedings of the 15th Conference of the European Chapter
  of the Association for Computational Linguistics (EACL)}, Valencia, Spain.

\bibitem[{Siegel and Castellan(1988)}]{Siegel/Castellan:88}
Sidney Siegel and N.~John Castellan. 1988.
\newblock \emph{Nonparametric Statistics for the Behavioral Sciences}.
\newblock McGraw-Hill, Boston, MA.

\bibitem[{Snow et~al.(2006)Snow, Jurafsky, and Ng}]{Snow:2006}
Rion Snow, Daniel Jurafsky, and Andrew~Y. Ng. 2006.
\newblock Semantic taxonomy induction from heterogenous evidence.
\newblock In \emph{Proceedings of the 21st Annual Meeting of the Association
  for Computational Linguistics (ACL)}, pages 801--808, Sydney, Australia.

\bibitem[{Sucameli(2015)}]{Sucameli15}
Irene Sucameli. 2015.
\newblock {A}nalisi computazionale delle relazioni semantiche: {U}no studio
  della lingua italiana.
\newblock B.s. thesis, University of Pisa.

\bibitem[{Tuan et~al.(2016)Tuan, Tay, Hui, and Ng}]{Tuan:16}
Luu~Anh Tuan, Yi~Tay, Siu~Cheung Hui, and See~Kiong Ng. 2016.
\newblock Learning term embeddings for taxonomic relation identification using
  dynamic weighting neural network.
\newblock In \emph{Proceedings of the 2016 Conference on Empirical Methods in
  Natural Language Processing (EMNLP)}, pages 403--413, Austin, Texas.

\bibitem[{Turney and Pantel(2010)}]{Turney/Pantel:10}
Peter~D. Turney and Patrick Pantel. 2010.
\newblock From {F}requency to {M}eaning: {V}ector {S}pace {M}odels of
  {S}emantics.
\newblock \emph{Journal of Artificial Intelligence Research}, 37:141--188.

\bibitem[{Vendrov et~al.(2016)Vendrov, Kiros, Fidler, and Urtasun}]{Vendrov:16}
Ivan Vendrov, Ryan Kiros, Sanja Fidler, and Raquel Urtasun. 2016.
\newblock Order-embeddings of images and language.
\newblock In \emph{Proceedings of the 4th International Conference on Learning
  Representations (ICLR)}, San Juan, Puerto Rico.

\bibitem[{Vilnis and McCallum(2015)}]{Vilnis/McCallum:15}
Luke Vilnis and Andrew McCallum. 2015.
\newblock Word representations via gaussian embedding.
\newblock In \emph{Proceedings of the 3rd International Conference on Learning
  Representations (ICLR)}, California, USA.

\bibitem[{Vuli{\'{c}} et~al.(2016)Vuli{\'{c}}, Gerz, Kiela, Hill, and
  Korhonen}]{Vulic:16}
Ivan Vuli{\'{c}}, Daniela Gerz, Douwe Kiela, Felix Hill, and Anna Korhonen.
  2016.
\newblock Hyperlex: {A} large-scale evaluation of graded lexical entailment.
\newblock \emph{arXiv}.

\bibitem[{Weeds et~al.(2014)Weeds, Clarke, Reffin, Weir, and Keller}]{Weeds:14}
Julie Weeds, Daoud Clarke, Jeremy Reffin, David~J. Weir, and Bill Keller. 2014.
\newblock Learning to distinguish hypernyms and co-hyponyms.
\newblock In \emph{Proceedings of the 25th International Conference on
  Computational Linguistics (COLING)}, pages 2249--2259, Dublin, Ireland.

\bibitem[{Weeds and Weir(2003)}]{Weeds/Weir:03}
Julie Weeds and David Weir. 2003.
\newblock A general framework for distributional similarity.
\newblock In \emph{Proceedings of the Conference on Empirical Methods in
  Natural Language Processing (EMNLP)}, pages 81--88, Stroudsburg, PA, USA.

\bibitem[{Weeds et~al.(2004)Weeds, Weir, and McCarthy}]{Weeds:04}
Julie Weeds, David Weir, and Diana McCarthy. 2004.
\newblock Characterising measures of lexical distributional similarity.
\newblock In \emph{Proceedings of the 20th International Conference on
  Computational Linguistics (COLING)}, pages 1015–--1021, Geneva,
  Switzerland.

\bibitem[{Wu and Palmer(1994)}]{Wu/Palmer:94}
Zhibiao Wu and Martha Palmer. 1994.
\newblock Verbs semantics and lexical selection.
\newblock In \emph{Proceedings of the 32nd Annual Meeting on Association for
  Computational Linguistics (ACL)}, pages 133--138, Las Cruces, New Mexico.

\bibitem[{Yu et~al.(2015)Yu, Wang, Lin, and Wang}]{Yu:15}
Zheng Yu, Haixun Wang, Xuemin Lin, and Min Wang. 2015.
\newblock Learning term embeddings for hypernymy identification.
\newblock In \emph{Proceedings of the 24th International Conference on
  Artificial Intelligence (IJCAI)}, pages 1390--1397, Buenos Aires, Argentina.

\bibitem[{Zhitomirsky-Geffet and Dagan(2009)}]{Geffet/Dagan:09}
Maayan Zhitomirsky-Geffet and Ido Dagan. 2009.
\newblock Bootstrapping distributional feature vector quality.
\newblock \emph{Computational Linguistics}, 35(3):435--461.

\end{thebibliography}
\bibliographystyle{emnlp_natbib}

\end{document}